\pdfoutput=1
\documentclass{article}

\usepackage[T1]{fontenc}
\PassOptionsToPackage{table}{xcolor}
\usepackage{harnessflow_arxiv,times}

\usepackage{amsmath,amsfonts,bm}

\def\eqref#1{equation~\ref{#1}}

\def\1{\bm{1}}

\DeclareMathAlphabet{\mathsfit}{\encodingdefault}{\sfdefault}{m}{sl}
\SetMathAlphabet{\mathsfit}{bold}{\encodingdefault}{\sfdefault}{bx}{n}

\usepackage{amsmath}
\usepackage{amssymb}
\usepackage{algorithm}
\usepackage{algpseudocode}
\usepackage{booktabs}
\usepackage{graphicx}
\usepackage{subcaption}
\usepackage{bm}
\usepackage{float}
\usepackage{array}
\usepackage{needspace}
\usepackage{hyperref}
\usepackage{url}

\hypersetup{
  hidelinks,
  pdftitle={HarnessFlow: MCTS-Guided Harness Optimization for LLM Agents},
  pdfauthor={Zhi Hong, Qianlong Wang, Hongchen Wei, Aidan Wu, Daohong Ying, Mingze Kong, Bei Liu, Chong Luo},
  pdfsubject={Preprint},
  pdfkeywords={LLM agents, harness optimization, Monte Carlo Tree Search}
}

\title{HarnessFlow: MCTS-Guided\\
Harness Optimization for LLM Agents}

\author{%
  Zhi Hong\textsuperscript{1,$\ddagger$}\quad
  Qianlong Wang\textsuperscript{2}\quad
  Hongchen Wei\textsuperscript{3,$\ddagger$}\quad
  Aidan Wu\textsuperscript{4,$\ddagger$}\\[0.4ex]
  Daohong Ying\textsuperscript{5,$\ddagger$}\quad
  Mingze Kong\textsuperscript{6}\quad
  Bei Liu\textsuperscript{7,$\ast$}\quad
  Chong Luo\textsuperscript{7}\\[1ex]
  {\small\textsuperscript{1}South China University of Technology\quad
   \textsuperscript{2}Fudan University\\[0.35ex]
   \textsuperscript{3}Wuhan University\quad
   \textsuperscript{4}Cornell University\quad
   \textsuperscript{5}University of California, San Diego\\[0.35ex]
   \textsuperscript{6}The Chinese University of Hong Kong, Shenzhen\quad
   \textsuperscript{7}Microsoft Research Asia}\\[0.6ex]
  {\small$^\ast$ Corresponding author.\quad
   $^\ddagger$ Work done during an internship at MSRA.}%
}

\begin{document}

\maketitle

\begin{abstract}
The performance of large language model (LLM) agents depends on both the
underlying model and the harness that governs interaction with tools
and the environment. Yet harness design remains largely manual, and
automating it requires navigating an open-ended search space through costly
end-to-end evaluations. We introduce \textbf{\textsc{HarnessFlow}}, a framework for
optimizing complete executable harnesses under a limited candidate-evaluation
budget, with the model and task environment held fixed.
\textsc{HarnessFlow} organizes candidates and their evaluation histories in
a refinement tree, using a Monte Carlo Tree Search (MCTS)-based policy to
balance exploitation and exploration. A coding agent refines selected
harnesses using archived code, execution feedback, and a library of reusable
harness mechanisms. With $15$ refinement attempts per benchmark,
\textsc{HarnessFlow} achieves the highest mean test scores among the compared
methods on six benchmarks spanning agentic tasks, coding, and web
interaction. The optimized harnesses achieve an average score of $69.48\%$,
outperforming the initial harnesses and Meta-Harness by $10.89$ and $3.95$
percentage points, respectively, while incurring lower average per-task
execution costs than hand-engineered baselines.
These results suggest that structured harness search offers a promising
route toward automating the design of capable and efficient LLM agents.
\end{abstract}

\section{Introduction}
\label{sec:introduction}

The performance of LLM agents depends not only on the underlying model but also
on the harness, the executable system that governs access to information,
maintains state, invokes tools, and coordinates execution~\citep{yang2024swe,shang2025agentsquare,khattab2024dspy,hu2025automated}. Prior studies of
context management and multi-agent systems show that these mechanisms can
substantially affect agent
behavior~\citep{packer2023memgpt,park2023generative},
highlighting harness design as a practical means of improving agent performance
without modifying model weights. However, harness design remains largely manual
and expertise-intensive, making effective designs difficult to adapt and reuse
across tasks and domains~\citep{zhang2025aflow}.

Recent work has sought to automate harness design and refinement to reduce
reliance on manual
engineering~\citep{zhuge2024language,yuan2025evoagent,zhang2024g,zhang2025multi,yue2025masrouter,hong2026maspob}.
Despite this progress, efficient optimization remains challenging: the harness
design space is open-ended, and evaluating each candidate requires costly
end-to-end agent execution. HARBOR~\citep{sengupta2026harbor},
PRISM~\citep{ruan2026beyond}, and Self-Harness~\citep{zhang2026self} refine
harnesses through predefined configurations or bounded edits.
Meta-Harness~\citep{lee2026meta} permits full-harness code refinement,
while AutoSaddler~\citep{park2026autosaddler} organizes durable updates from
agent execution traces. DarwinX~\citep{zhang2026darwinx} explores harness
populations through selection and preservation checks. These approaches
motivate the study of search organization under a limited evaluation budget,
leading to a central question:
\emph{How can we systematically search over complete executable harnesses while
making effective use of a limited candidate-evaluation budget?}

\begin{figure}[t]
\centering
\begin{subfigure}[b]{0.58\textwidth}
    \centering
    \includegraphics[width=\linewidth]{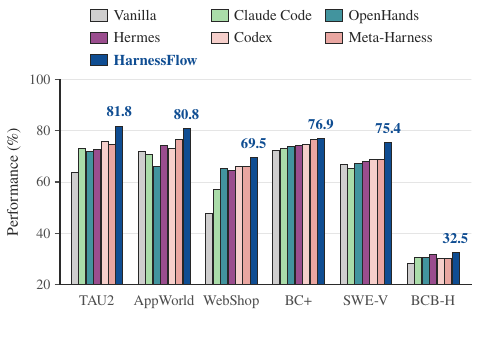}
    \caption{Performance comparison across benchmarks.}
    \label{fig:main-a}
\end{subfigure}
\hfill
\begin{subfigure}[b]{0.40\textwidth}
    \centering
    \includegraphics[width=\linewidth]{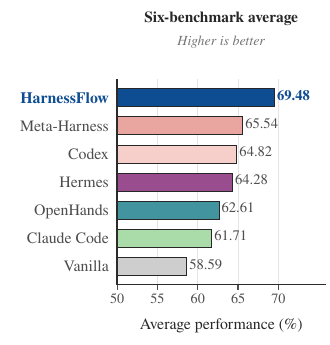}
    \caption{Overall performance comparison.}
    \label{fig:main-b}
\end{subfigure}
\caption{Performance across six benchmarks. (a) Scores on individual
benchmarks; labels indicate the best score in each group. (b) Ranking by
average score. \textsc{HarnessFlow} outperforms Vanilla and the strongest prior
method by $10.89$ and $3.95$ points on average, respectively.}
\label{fig:main}
\vspace{-12pt}
\end{figure}

\Needspace{3\baselineskip}
To address this challenge, we introduce \textbf{\textsc{HarnessFlow}}, a Monte Carlo
Tree Search (MCTS)-based framework~\citep{browne2012survey} designed to systematically explore and
discover optimal harnesses under a limited evaluation budget. First, it
represents each harness as a complete executable codebase that a coding agent
can inspect and modify in a filesystem workspace, enabling coordinated changes
across the entire harness~\citep{chen2026harnessx,lee2026meta}. It also provides the coding agent with a
\emph{Harness Component Library} of reusable mechanisms that can be adapted,
composed, or extended during refinement. Second, it reformulates harness
optimization as a tree-structured search problem over these executable
codebases: nodes represent harness implementations, and edges record code-level
modifications. The tree preserves earlier designs together with their
evaluation scores and execution traces, providing structured experience that
the coding agent can inspect and reuse across branches. Finally, it employs
an MCTS-based selection policy to determine which eligible candidate in this
tree to refine next, while the coding agent determines how to modify it. The
policy combines measured performance with prior expansion counts to balance
exploitation of promising designs with exploration of underexplored refinement
paths~\citep{auer2002finite,kocsis2006bandit}.

We conduct extensive evaluations on six established benchmarks spanning
agentic tasks (TAU2-bench and AppWorld), code generation
(BigCodeBench-Hard and SWE-bench Verified), and web interaction
(WebShop and BrowseComp-Plus). The results demonstrate that
\textsc{HarnessFlow} is \textbf{(i) high-performing}, achieving the highest
average test score of $69.48\%$ in the main comparison and improving over
Vanilla by $10.89$ percentage points; \textbf{(ii) cost-efficient}, running
at a lower average inference cost than every hand-engineered harness, at
$41.4\%$ of Codex's across the six benchmarks; and \textbf{(iii) sample-efficient},
attaining a mean best-so-far validation score of $71.5\%$ after three
candidate evaluations per benchmark and $74.2\%$ by round $15$.

\Needspace{5\baselineskip}
In summary, our key contributions are as follows:
\begin{itemize}
    \item We formulate full-harness optimization as a tree-structured search
    problem over complete executable codebases under a limited
    candidate-evaluation budget.
    \item We introduce \textsc{HarnessFlow}, an MCTS-based framework that
    uses a coding agent, tree-structured experience, and a Harness Component
    Library to generate full-harness refinements, while an explicit selection
    policy allocates the search budget across alternative designs.
    \item We conduct extensive evaluations on six benchmarks, demonstrating
    that \textsc{HarnessFlow} achieves strong task performance, cost-efficient
    inference, and sample-efficient harness optimization.
\end{itemize}

\begin{figure}[!t]
\centering
\includegraphics[width=\textwidth]{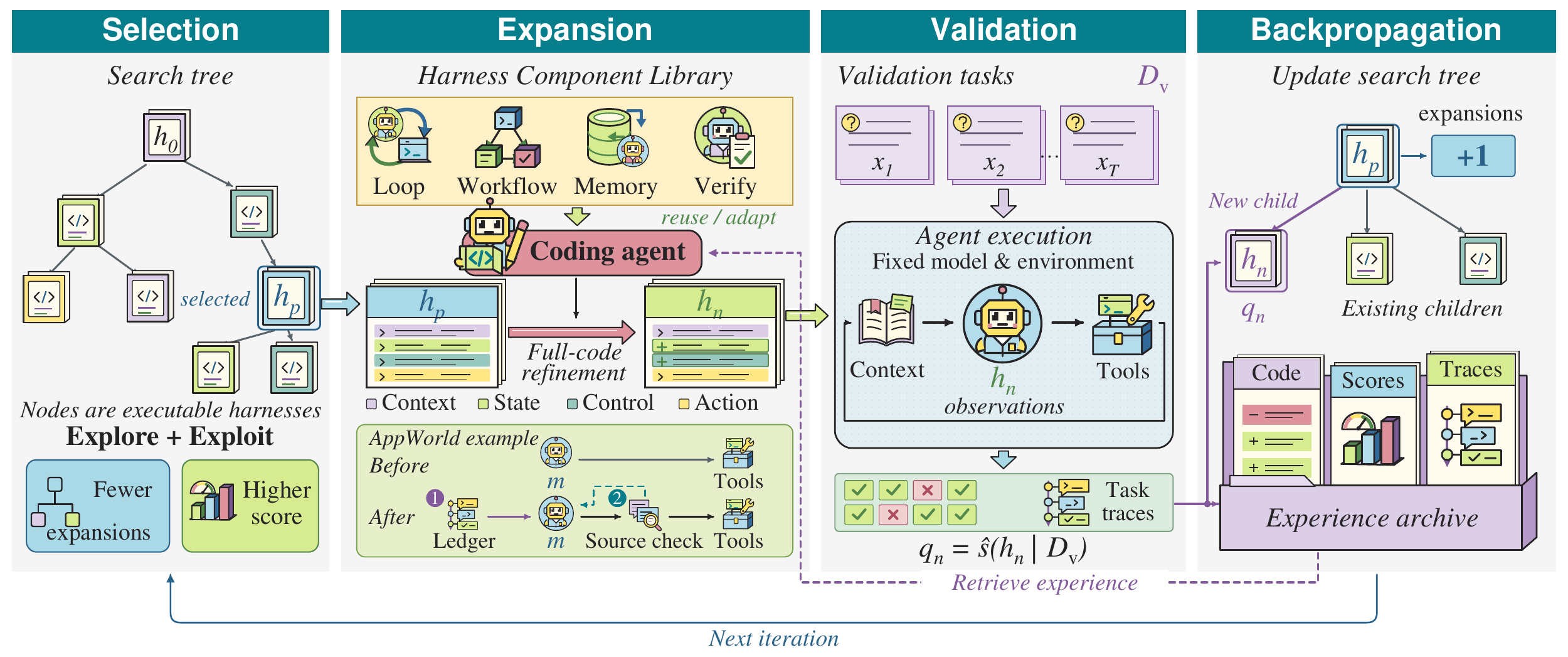}
\caption{Overview of \textsc{HarnessFlow}. Selection, expansion, validation,
and experience backpropagation organize full-harness search. A coding agent
uses reusable components and archived code, scores, and traces to refine
selected harnesses.}
\label{fig:framework}
\end{figure}

\section{Problem Setting}
\label{sec:problem}

\paragraph{Harness and Search Space.}
Given a fixed model $m$, task environment, and evaluation protocol, a
\emph{harness} $h$ is an executable program that governs how the model
interacts with tools and the environment. Let $\mathcal{H}$ denote the space
of harnesses compatible with this setting.
Refinements may modify mechanisms for context construction, state management,
tool use, and termination, as well as their composition and the overall
execution structure. For example, adding a verification step may require
retaining additional state and revising the stopping rule. The search
therefore operates over complete
executable harnesses rather than a predefined set of prompts, configurations,
or components.

\paragraph{Harness Evaluation and Optimization Objective.}
For a task set $\mathcal{D}=\{(x,y)\}$, $x$ denotes a task instance and $y$
the corresponding reference answer or success criterion. We write
$\operatorname{Agent}(m,h,x)$ for the end-to-end outcome produced by running
harness $h$ with model $m$ on $x$. Depending on the task, this outcome may be
a final response, a generated artifact, or the terminal state of an
interactive environment. A benchmark-specific metric $r(\cdot,\cdot)$
compares this outcome with $y$ and returns a normalized score in $[0,1]$.
We define harness performance as
\begin{equation}
    s(h\mid\mathcal{D})
    = \mathbb{E}_{(x,y)\sim\mathcal{D}}
      \left[r\!\left(\operatorname{Agent}(m,h,x),y\right)\right],
    \label{eq:performance}
\end{equation}
where tasks are sampled uniformly from $\mathcal{D}$. In practice, this score
is estimated by averaging repeated executions on the task set. It measures
end-to-end harness performance rather than individual model calls. The
resulting scores support candidate comparison, while execution traces provide
diagnostic evidence for subsequent refinements. Given a validation set
$\mathcal{D}_{\mathrm{v}}$, the ideal optimization objective is
\begin{equation}
    h^\star\in\arg\max_{h\in\mathcal{H}}
    s(h\mid\mathcal{D}_{\mathrm{v}}).
    \label{eq:objective}
\end{equation}

\paragraph{Budgeted Optimization Protocol.}
Evaluating a harness on $\mathcal{D}_{\mathrm{v}}$ is costly, as each task may
require multiple model and tool calls. We therefore consider a sequential
optimization protocol initialized from a harness $h_0$ and constrained by a
budget of $N$ refinement attempts. At each attempt, the optimizer draws on
accumulated validation scores and execution traces to select an existing
harness for refinement and propose a new candidate. The central challenge
is to allocate refinement
effort across alternative paths while further developing promising designs.
At the end of search, we select the harness with the highest observed score
on $\mathcal{D}_{\mathrm{v}}$ among the candidates produced during
optimization. The selected harness is then evaluated on a
disjoint held-out test set $\mathcal{D}_{\mathrm{t}}$, and the resulting score
is reported as its final performance.

\section{Methodology}
\label{sec:method}

\textsc{HarnessFlow} implements the protocol in Section~\ref{sec:problem} by
separating parent selection from code refinement. An MCTS-based policy
determines which harness to refine, while a coding agent determines how to
modify it. The policy allocates search effort using validation scores and
expansion counts; the coding agent uses prior execution experience and
general, reusable harness mechanisms to develop new candidates. Each
evaluation contributes evidence that can inform subsequent selection and
refinement. Figure~\ref{fig:framework} illustrates this cycle, and
Algorithm~\ref{alg:harnessflow} summarizes the procedure.

\subsection{Coding-Agent Full-Harness Refinement}
\label{sec:refinement}

\paragraph{Code-Represented Harnesses.}
Each candidate is represented by a complete executable codebase. Refinement
may add, remove, or jointly modify mechanisms for context construction,
state management, control flow, and tool use. This allows a single proposal
to coordinate changes across interacting mechanisms, supporting both local
improvements and broader architectural changes within the search space
defined in Section~\ref{sec:problem}.

\paragraph{Experience-Guided Code Refinement.}
The coding agent develops each candidate from a selected parent harness,
guided by accumulated search experience. We denote the experience archive
by $\mathcal{A}$; it contains prior implementations, code modifications,
validation feedback, task-level execution records, and refinement
relationships. The parent supplies the starting design, while the archive
provides empirical evidence about the effects of earlier changes. The agent
can selectively retrieve relevant code and feedback to compare designs,
diagnose failures, and identify refinements, without requiring the entire
search history to fit in its context.

\paragraph{Harness Component Library.}
The Harness Component Library $\mathcal{L}$ offers reusable code for context
construction, memory and state tracking, control flow, and tool interfacing
(Figure~\ref{fig:harness-library} in the appendix). It includes
observation compaction and skill catalogs; note stores, observation logs, and
request tracking; loop and workflow runners; and argument-schema checks with
heuristic tool-effect classification. The coding agent can inspect a read-only
snapshot and is instructed to copy adopted code and local dependencies into
the candidate, not import the snapshot at runtime. It may adapt or compose
these implementations or write new code. The library thus offers optional
implementation references rather than a fixed template or an exhaustive
search space; the experience archive supplies empirical feedback for refinement.

\subsection{MCTS-Guided Tree Search}
\label{sec:selection}

After $n$ refinement attempts, the search retains evaluated candidates
in a rooted tree $\mathcal{G}_n=(\mathcal{V}_n,\mathcal{E}_n)$. Each node
$u_i\in\mathcal{V}_n$ represents a harness $h_i$ together with its validation
score $q_i$ and execution records $\mathcal{R}_i$; the root $u_0$ represents
the initial harness $h_0$. An edge $(u_i,u_j)$ indicates that $h_j$ was
generated by refining $h_i$. Each evaluated harness and its measured score
remain unchanged as the tree grows, preserving earlier designs and alternative
refinement paths. Expansion counts and accumulated experience evolve
with the search. \textsc{HarnessFlow} adapts the MCTS
cycle~\citep{browne2012survey} through selection, expansion, evaluation, and
experience backpropagation.

\paragraph{Selection.}
At attempt $n$, the optimizer selects a parent from
$\mathcal{C}_n\subseteq\mathcal{V}_{n-1}$, comprising up to $K$ highest-scoring
nodes, where $K\geq1$ is the shortlist size. This restriction applies to
parent selection; the coding agent may still use code and evidence from
other archived candidates. Within $\mathcal{C}_n$, exploitation favors higher
validation scores:
\begin{equation}
    P_{\mathrm{exploit}}(i)
    =\frac{\exp(q_i/\tau)}
          {\sum_{u_j\in\mathcal{C}_n}\exp(q_j/\tau)},
    \label{eq:exploitation}
\end{equation}
for $u_i\in\mathcal{C}_n$, where the temperature $\tau>0$ controls sensitivity
to score differences. Smaller values concentrate probability on
higher-scoring candidates. Its setting is described in
Section~\ref{sec:setup}.

Exploration prioritizes underexplored candidates. Let $d_i$ denote the
number of direct children of node $u_i$ in the search tree. We define
\begin{equation}
    P_{\mathrm{explore}}(i)
    =\frac{(1+d_i)^{-\alpha}}
          {\sum_{u_j\in\mathcal{C}_n}(1+d_j)^{-\alpha}},
    \label{eq:exploration}
\end{equation}
where $\alpha\geq0$ controls the strength of this exploration preference.
The count measures direct expansions rather than the size of the full
descendant subtree.

The parent-selection policy combines the two distributions:
\begin{equation}
    \pi_n(i)=(1-\varepsilon)P_{\mathrm{exploit}}(i)
              +\varepsilon P_{\mathrm{explore}}(i),
    \qquad p_n\sim\pi_n,
    \label{eq:selection}
\end{equation}
where $\varepsilon\in[0,1]$ balances exploitation and exploration. The first
term favors high-performing harnesses, while the second directs refinement
toward underexplored candidates. Refinement effort therefore depends on
both the quality of a harness and how extensively it has already been
developed.

\paragraph{Expansion.}
Given the selected parent $h_{p_n}$, the coding agent proposes a new harness
$h_n$ and an edit record $\Delta_n$ using the refinement mechanism of
Section~\ref{sec:refinement}:
\begin{equation}
    (h_n,\Delta_n)\gets
    \operatorname{Refine}(h_{p_n};\mathcal{A},\mathcal{L}).
    \label{eq:refinement}
\end{equation}
The selected parent defines the refinement relationship, while the archive
allows relevant code and evidence from other branches to inform the proposal.

\paragraph{Evaluation.}
Evaluation provides an observed validation score $q_n$ and task-level
execution records $\mathcal{R}_n$. The score estimates
$s(h_n\mid\mathcal{D}_{\mathrm{v}})$ from $R$ repeated executions under the
fixed model and task environment. The score supports candidate comparison,
while the execution records guide subsequent refinement.

\paragraph{Experience Backpropagation.}
Each new tree node $u_n$ is linked to its parent by edge $(u_{p_n},u_n)$,
and $d_{p_n}$ is incremented. The candidate's code changes $\Delta_n$,
validation score,
parent-relative performance change, and task-level records extend
$\mathcal{A}$ together with the refinement relationship. Both
improving and non-improving refinements contribute to this experience,
allowing later proposals to build on useful changes and reconsider
unsuccessful directions. Backpropagation therefore updates refinement
relationships and accumulated evidence while preserving each harness's
directly measured score $q_i$. After $N$ attempts, final selection follows
the validation criterion in Section~\ref{sec:problem}.

\begin{algorithm}[t]
    \caption{\textsc{HarnessFlow}: MCTS-guided full-harness optimization}
    \label{alg:harnessflow}
    \begin{algorithmic}[1]
        \Require Initial harness $h_0$; component library $\mathcal{L}$;
        validation set $\mathcal{D}_{\mathrm{v}}$; evaluation repetitions $R$;
        \mbox{refinement budget $N>0$};
        selection parameters $K,\varepsilon,\alpha,\tau$
        \Ensure Harness $h^\star$ with the highest observed validation score
        \State $(\_,q_0,\mathcal{R}_0)\gets
        \Call{Evaluate}{h_0,\mathcal{D}_{\mathrm{v}},R}$
        \State Initialize tree $\mathcal{G}$ with root
        $u_0=(h_0,q_0,\mathcal{R}_0)$; set $d_0\gets0$
        \State Initialize experience archive
        $\mathcal{A}\gets\{(h_0,q_0,\mathcal{R}_0)\}$
        \For{$n=1,\ldots,N$}
            \Statex \textbf{Selection:}
            \State $\mathcal{C}_n\gets$ up to $K$ highest-scoring nodes in $\mathcal{G}$
            \State Compute $\pi_n$ using Equation~\ref{eq:selection};
            sample $p_n\sim\pi_n$
            \Statex \textbf{Expansion:}
            \State $(h_n,\Delta_n)\gets
            \Call{Refine}{h_{p_n};\mathcal{A},\mathcal{L}}$
            \Statex \textbf{Evaluation:}
            \State $(v_n,q_n,\mathcal{R}_n)\gets
            \Call{Evaluate}{h_n,\mathcal{D}_{\mathrm{v}},R}$
            \Statex \textbf{Experience Backpropagation:}
            \State Record $h_n$, parent $p_n$, edits $\Delta_n$,
            and available evaluation feedback in $\mathcal{A}$
            \If{$v_n$ is valid} \Comment{See Appendix~\ref{app:config}}
                \State Add $u_n=(h_n,q_n,\mathcal{R}_n)$ and
                edge $(u_{p_n},u_n)$ to $\mathcal{G}$
                \State $d_n\gets0$; $d_{p_n}\gets d_{p_n}+1$
                \State Store the parent-relative score change $q_n-q_{p_n}$ in $\mathcal{A}$
            \EndIf
        \EndFor
        \State Select $u_{i^\star}\in\arg\max_{u_i\in\mathcal{V}(\mathcal{G})}q_i$
        \State \Return $h_{i^\star}$
    \end{algorithmic}
\end{algorithm}

\section{Experiments}
\label{sec:experiments}

\subsection{Experimental Setup}
\label{sec:setup}

\paragraph{Benchmarks.}
We evaluate on six established benchmarks spanning agentic tasks, code
generation, and web interaction:
TAU2-bench~\citep{barres2025tau}, AppWorld~\citep{trivedi2024appworld},
SWE-bench Verified~\citep{jimenez2024swe}, BigCodeBench-Hard~\citep{zhuo2025bigcodebench},
WebShop~\citep{yao2022webshop}, and BrowseComp-Plus~\citep{chen2026browsecomp}.
Each benchmark is split once, with a fixed seed, into a validation set used
during search and a disjoint test set used only for the final evaluation
(92/186, 105/210, 100/200, 100/200, 100/200, and 100/200 tasks, respectively).
The test set is never observed by the optimizer.

\paragraph{Baselines.}
We compare against three categories of methods.
\textit{(i)} \textbf{Vanilla}, the initial harness from which search starts: the
benchmark's own reference agent implementation, which is also the root node of
the search tree.
\textit{(ii)} \textbf{Production agent harnesses} that are engineered by hand
and widely used: Claude Code~\citep{claudecode}, Codex~\citep{codexcli},
OpenHands~\citep{wang2025openhands}, and Hermes~\citep{hermesagent}.
We evaluate the official implementations at fixed versions, using the same
backbone model, access to benchmark tools, and per-task resource limits as
\textsc{HarnessFlow}.
Appendix~\ref{app:harness0} documents the
origin of every initial harness, and Appendix~\ref{app:agents} the exact releases and
invocation of the production agents.
\textit{(iii)} \textbf{Meta-Harness}~\citep{lee2026meta},
a harness-optimization method that, like ours, permits full-code modification of
a complete harness, but refines proposals iteratively from execution feedback
rather than maintaining a persistent search tree. It uses the same initial
harness, coding agent, benchmark, candidate budget, and evaluation protocol.

\paragraph{Metrics.}
Following prior LLM agent work~\citep{zhou2023language,zhang2026self},
we report each benchmark's official metric in percent: task success rate for
TAU2-bench and AppWorld; task score for WebShop; resolution rate for SWE-bench
Verified;
Pass@1 for BigCodeBench-Hard; and answer accuracy for BrowseComp-Plus.

\paragraph{Implementation Details.}
Unless stated otherwise, all agents execute with GPT-5.4-mini~\citep{openai2026gpt54mini} at medium
reasoning effort as the backbone model, and optimization uses a budget of $N=15$
candidate attempts. Every method is executed $R=5$ times over each split,
during search and at test time alike, and tables report the mean of those five
executions $\pm$ their sample standard deviation. For the policy in
Section~\ref{sec:selection}, we use $\varepsilon=0.35$, $\alpha=1$, and
automatically determined $\tau$. At each attempt, $\mathcal{C}_n$ contains
up to $K=5$ highest-scoring nodes, with ties broken by creation order.\footnote{We set
$\tau=\operatorname{clip}(\max_{i\in J}\sqrt{q_i(1-q_i)/M_i},0.005,0.100)$, one
standard error of the validation score across the shortlisted nodes, where
$M_i$ is the number of evaluation cases for node $i$ and $J$ contains
shortlisted nodes with $M_i>0$ and $0<q_i<1$; if $J$ is empty, $\tau=0.08$.
Tying the temperature to the evaluation's own noise keeps the policy free of a
per-benchmark constant.} The general-purpose coding agent that performs
$\operatorname{Refine}$ refines the selected parent using the search history;
it never sees test tasks. Each run uses a fixed snapshot of the execution
runtime, so search, baselines, and final evaluation execute identical runtime
code. Further settings are
given in Appendix~\ref{app:config}.

\begin{table}[t]
\centering
\caption{Performance comparison across six widely used benchmarks. We report mean score (\%) $\pm$ standard deviation over five runs. All methods use the same backbone model, tool access, and per-task limits. Bold marks the highest mean score for each benchmark.}
\label{tab:main}
\small
\setlength{\tabcolsep}{1.75pt}
\renewcommand{\arraystretch}{1.15}
\begin{tabular}{@{}l cc cc cc c@{}}
\toprule
& \multicolumn{2}{c}{\textbf{Agentic}} & \multicolumn{2}{c}{\textbf{Web}} & \multicolumn{2}{c}{\textbf{Code}} & \\
\cmidrule(lr){2-3} \cmidrule(lr){4-5} \cmidrule(lr){6-7}
\textbf{Method} & \textbf{TAU2} & \textbf{AppWorld} & \textbf{WebShop} & \textbf{BC+} & \textbf{SWE-V} & \textbf{BCB-H} & \textbf{Avg.} \\
\midrule
Vanilla & 63.98{\scriptsize$\pm$2.15} & 72.00{\scriptsize$\pm$1.86} & 47.79{\scriptsize$\pm$1.14} & 72.40{\scriptsize$\pm$1.85} & 67.10{\scriptsize$\pm$0.89} & 28.30{\scriptsize$\pm$0.76} & 58.59 \\
\midrule
Claude Code~\citep{claudecode} & 73.12{\scriptsize$\pm$1.14} & 70.86{\scriptsize$\pm$2.00} & 56.98{\scriptsize$\pm$1.19} & 73.20{\scriptsize$\pm$1.72} & 65.30{\scriptsize$\pm$1.30} & 30.80{\scriptsize$\pm$2.61} & 61.71 \\
OpenHands~\citep{wang2025openhands} & 71.83{\scriptsize$\pm$1.59} & 66.29{\scriptsize$\pm$1.32} & 65.54{\scriptsize$\pm$1.77} & 74.00{\scriptsize$\pm$1.97} & 67.20{\scriptsize$\pm$1.64} & 30.80{\scriptsize$\pm$2.86} & 62.61 \\
Hermes~\citep{hermesagent} & 72.90{\scriptsize$\pm$2.17} & 74.19{\scriptsize$\pm$1.03} & 64.50{\scriptsize$\pm$1.34} & 74.30{\scriptsize$\pm$1.30} & 68.10{\scriptsize$\pm$2.53} & 31.70{\scriptsize$\pm$1.15} & 64.28 \\
Codex~\citep{codexcli} & 75.91{\scriptsize$\pm$2.98} & 73.24{\scriptsize$\pm$1.97} & 66.15{\scriptsize$\pm$1.49} & 74.70{\scriptsize$\pm$2.36} & 68.70{\scriptsize$\pm$2.11} & 30.20{\scriptsize$\pm$1.10} & 64.82 \\
\midrule
Meta-Harness~\citep{lee2026meta} & 74.62{\scriptsize$\pm$2.36} & 76.76{\scriptsize$\pm$2.94} & 66.33{\scriptsize$\pm$0.61} & 76.60{\scriptsize$\pm$1.02} & 68.80{\scriptsize$\pm$1.15} & 30.10{\scriptsize$\pm$1.66} & 65.53 \\
\rowcolor{gray!15}
\textbf{\textsc{HarnessFlow}} & \textbf{81.83}{\scriptsize$\pm$2.39} & \textbf{80.76}{\scriptsize$\pm$1.33} & \textbf{69.49}{\scriptsize$\pm$0.88} & \textbf{76.90}{\scriptsize$\pm$1.14} & \textbf{75.40}{\scriptsize$\pm$1.19} & \textbf{32.50}{\scriptsize$\pm$1.77} & \textbf{69.48} \\
\bottomrule
\end{tabular}
\end{table}

\subsection{Main Results and Analysis}

\paragraph{Main Results.}
\textsc{HarnessFlow} achieves the highest mean test score on all six
benchmarks, averaging $69.48\%$ (Table~\ref{tab:main}, Figure~\ref{fig:main}).
It exceeds Vanilla, Codex, and Meta-Harness by $10.89$, $4.66$, and $3.95$
percentage points on average, respectively. Gains over Vanilla are largest
on WebShop ($+21.70$) and TAU2-bench ($+17.85$). Against Meta-Harness under
the same $15$-candidate budget, the largest gains are on TAU2-bench
($+7.21$) and SWE-bench Verified ($+6.60$).

\begin{figure}[t]
\centering
\begin{subfigure}[b]{0.49\textwidth}
    \centering
    \includegraphics[width=\linewidth]{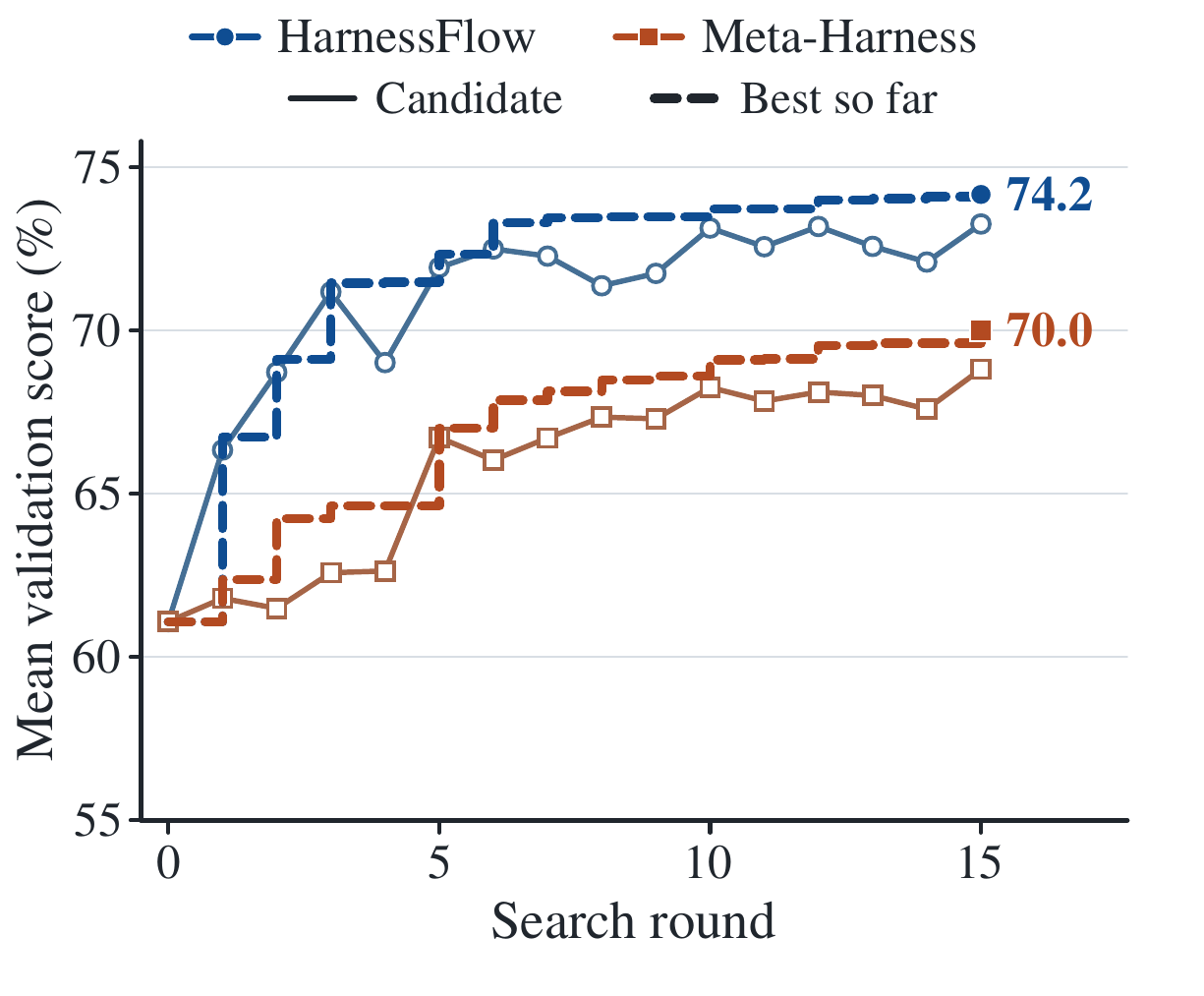}
    \caption{Search efficiency on the validation set.}
    \label{fig:curves}
\end{subfigure}
\hfill
\begin{subfigure}[b]{0.49\textwidth}
    \centering
    \includegraphics[width=\linewidth]{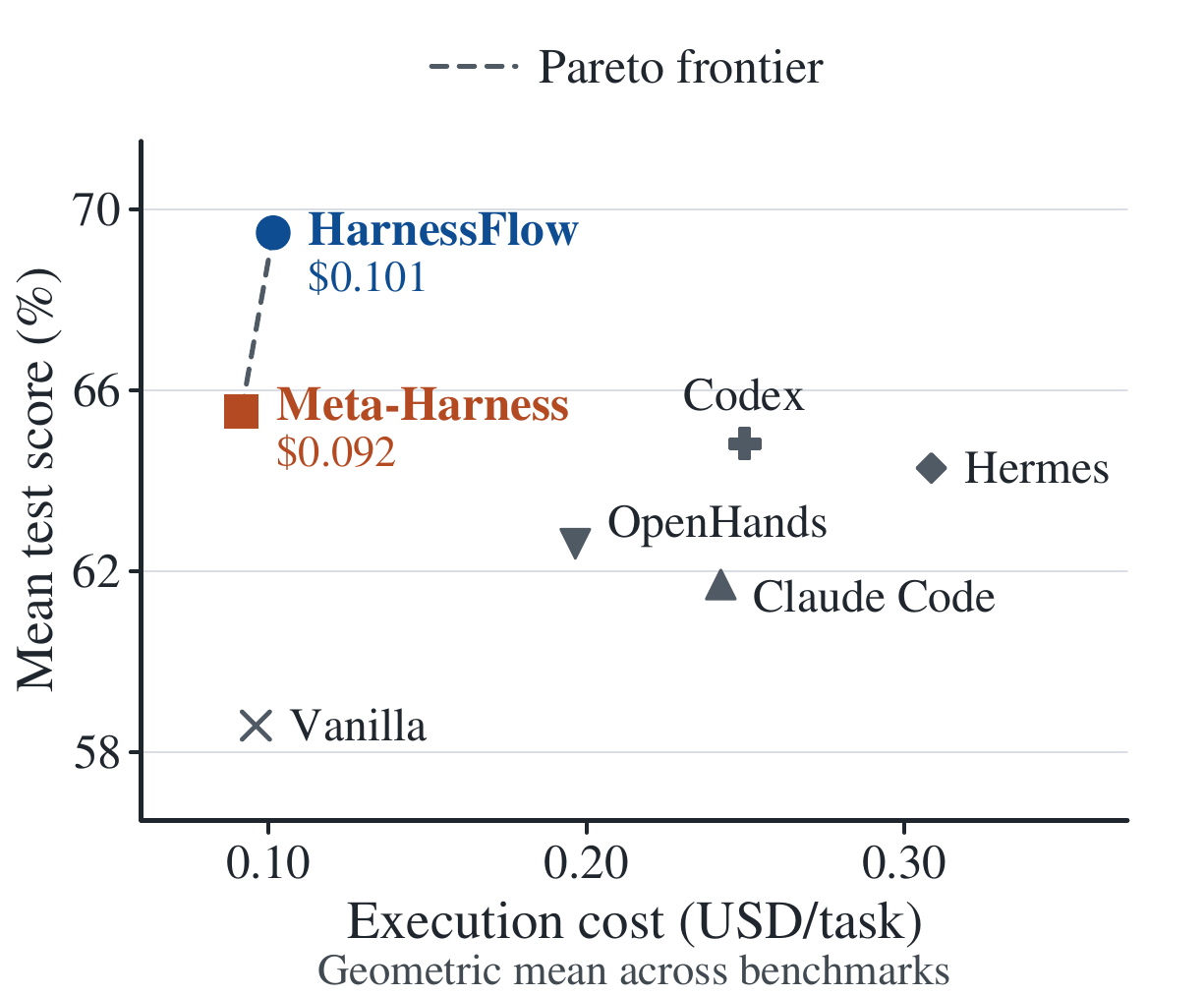}
    \caption{Test performance versus execution cost.}
    \label{fig:pareto}
\end{subfigure}
\caption{Search efficiency and execution cost across six benchmarks.
(a) Mean candidate validation scores (solid) and mean per-benchmark best-so-far
scores (dashed), from a shared round-$0$ harness. (b) Mean test score versus geometric-mean
execution cost per task, excluding search cost; dashed segment: Pareto
frontier. Per-benchmark results: Appendices~\ref{app:search}
and~\ref{app:cost}.}
\label{fig:efficiency}
\end{figure}

\paragraph{Search Efficiency.}
\textsc{HarnessFlow} exhibits strong \emph{sample efficiency} under a
limited candidate-evaluation budget
(Figure~\ref{fig:curves}). After just three candidate evaluations per
benchmark, its mean best validation score reaches $71.5\%$. Continued search raises the mean best
validation score to $74.2\%$ by round $15$. On four of the six benchmarks,
the final best is first reached at round $10$ or later. These candidates
refine earlier retained harnesses rather than the immediately preceding
candidate, illustrating how the tree supports revisiting promising designs
throughout search. Per-benchmark curves and lineage examples are provided in
Appendices~\ref{app:search} and~\ref{app:search-evolution}.

\Needspace{6\baselineskip}
\paragraph{Execution Cost.}
Figure~\ref{fig:pareto} compares test scores against the geometric mean of
per-task execution costs across benchmarks, reducing the influence of the
costlier BrowseComp-Plus tasks. \textsc{HarnessFlow} and Meta-Harness form the
score--cost Pareto frontier at $\$0.101$ and $\$0.092$, respectively;
hand-engineered harnesses cost two to three times as much at lower scores.
These estimates exclude one-time harness-search costs; per-benchmark plots
are given in Appendix~\ref{app:cost}.

\begin{figure}[!t]
\centering
\includegraphics[width=\textwidth]{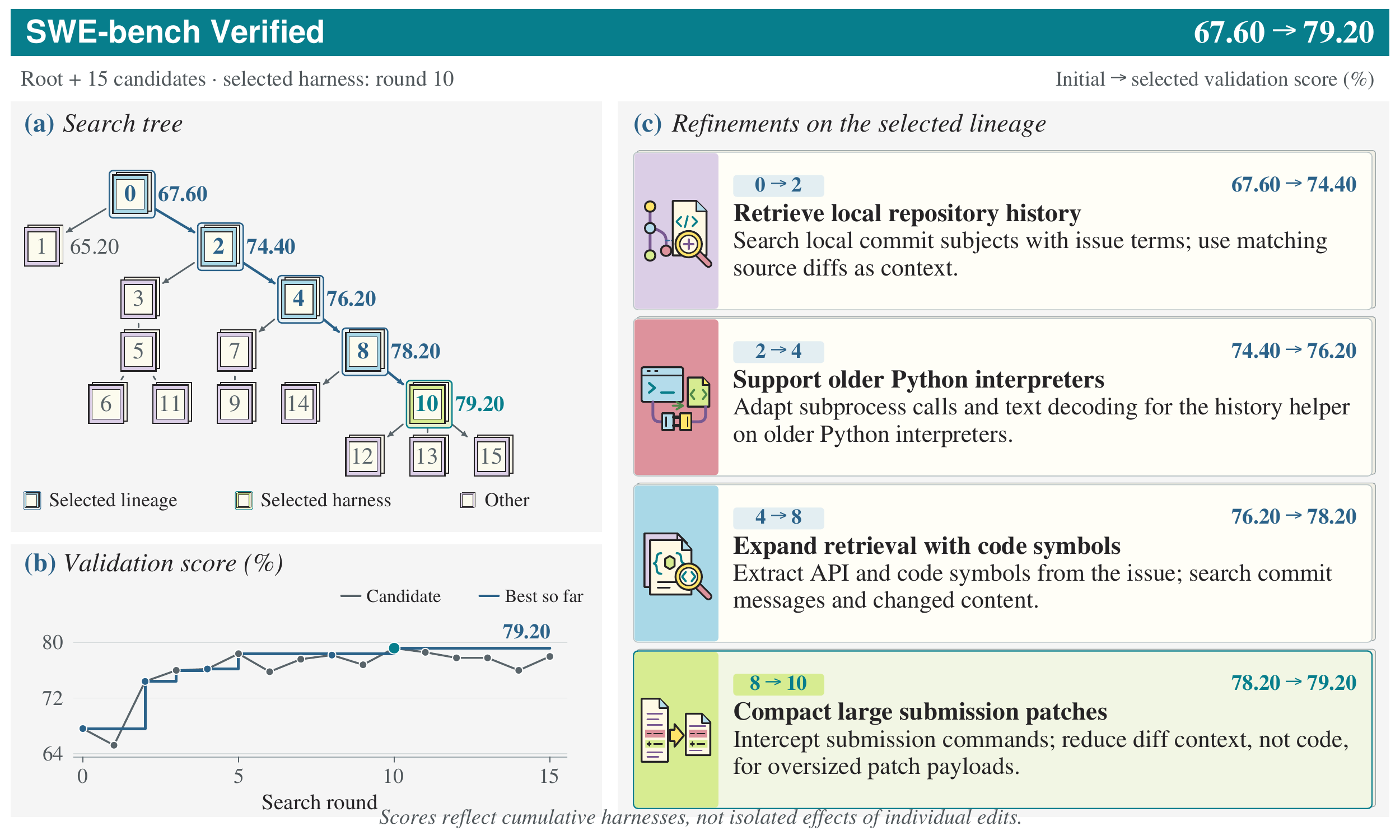}
\caption{Harness evolution on SWE-bench Verified. Code-level refinements
along the selected lineage raise the validation score from $67.60\%$ to
$79.20\%$. Parent--child labels show actual search edges; scores describe
cumulative harness designs, not isolated effects of individual edits.}
\label{fig:evolution-swe}
\end{figure}

\paragraph{Harness Evolution.}
Figure~\ref{fig:evolution-swe} makes the search process concrete on
SWE-bench Verified. Code-level refinements along the selected lineage
$0\!\to\!2\!\to\!4\!\to\!8\!\to\!10$ raise its validation score from
$67.60\%$ to $79.20\%$: repository-history retrieval, interpreter
compatibility, symbol-level retrieval, and compaction of large submission
patches are added in turn, each building on a retained parent. Later attempts
refine both the best candidate and earlier branches, and the tree keeps every
evaluated design available, so the final selection is the best harness
observed over the whole search. A corresponding harness-evolution example on
TAU2-bench is shown in Figure~\ref{fig:evolution-tau2} in
Appendix~\ref{app:search-evolution}.

\subsection{Ablation Study}
\label{sec:ablation}

\begin{table}[t]
\centering
\captionsetup{position=top,skip=4pt,justification=raggedright,singlelinecheck=false}
\begin{minipage}[t]{0.545\textwidth}
\vspace{0pt}
\caption{Library and initialization ablations. Search is repeated without
the component library or from the OpenHands agent, with other settings
unchanged. Scores are test means (\%) $\pm$ standard deviations.}
\label{tab:ablation}
\end{minipage}\hfill
\begin{minipage}[t]{0.435\textwidth}
\vspace{0pt}
\caption{Exploration coefficient ablation. We compare $\varepsilon=0$
(exploitation), $0.35$ (default), and $1$ (exploration), holding all other
settings fixed.}
\label{tab:epsilon}
\end{minipage}
\par\nointerlineskip
\begin{minipage}[t]{0.545\textwidth}
\vspace{0pt}
\scriptsize
\setlength{\tabcolsep}{2.6pt}
\renewcommand{\arraystretch}{1.15}
\begin{tabular*}{\linewidth}{@{\extracolsep{\fill}}lcccc@{}}
\toprule
\textbf{Method} & \textbf{TAU2} & \textbf{AppWorld} & \textbf{BCB-H} & \textbf{Avg.} \\
\midrule
Vanilla & 63.98{\tiny$\pm$2.15} & 72.00{\tiny$\pm$1.86} & 28.30{\tiny$\pm$0.76} & 54.76 \\
\textsc{HarnessFlow} & \textbf{81.83}{\tiny$\pm$2.39} & \textbf{80.76}{\tiny$\pm$1.33} & \textbf{32.50}{\tiny$\pm$1.77} & \textbf{65.03} \\
\quad without library & 77.64{\tiny$\pm$1.70} & 80.63{\tiny$\pm$0.99} & 31.90{\tiny$\pm$1.67} & 63.39 \\
\quad from OpenHands agent & 76.02{\tiny$\pm$2.42} & 79.62{\tiny$\pm$0.92} & 32.20{\tiny$\pm$1.96} & 62.61 \\
\bottomrule
\end{tabular*}
\end{minipage}\hfill
\begin{minipage}[t]{0.435\textwidth}
\vspace{0pt}
\scriptsize
\setlength{\tabcolsep}{3.4pt}
\renewcommand{\arraystretch}{1.15}
\begin{tabular*}{\linewidth}{@{\extracolsep{\fill}}lccc@{}}
\toprule
\textbf{Benchmark} & $\boldsymbol{\varepsilon=0}$ & $\boldsymbol{\varepsilon=0.35}$ & $\boldsymbol{\varepsilon=1}$ \\
\midrule
TAU2 & 74.19{\tiny$\pm$2.77} & \textbf{81.83}{\tiny$\pm$2.39} & 72.80{\tiny$\pm$2.10} \\
AppWorld & 78.76{\tiny$\pm$2.47} & \textbf{80.76}{\tiny$\pm$1.33} & 79.81{\tiny$\pm$0.80} \\
SWE-V & 68.83{\tiny$\pm$2.57} & \textbf{75.40}{\tiny$\pm$1.19} & 68.33{\tiny$\pm$0.29} \\
Avg. & 73.93 & \textbf{79.33} & 73.65 \\
\bottomrule
\end{tabular*}
\end{minipage}
\end{table}

\paragraph{Library and Initialization.}
Removing the Harness Component Library lowers the three-benchmark average
from $65.03$ to $63.39$ (Table~\ref{tab:ablation}), primarily through
TAU2-bench ($81.83\rightarrow77.64$), with small changes on AppWorld and
BigCodeBench-Hard. Without the library, search still improves all three
benchmarks over Vanilla by $8.63$ points on average. Reusable components
complement the search procedure, with the clearest gains on TAU2-bench.
Starting from OpenHands likewise
improves all three benchmarks, reaching an average of $62.61$. The gains
therefore extend to another agent architecture.

\paragraph{Exploration--Exploitation Balance.}
To examine the contribution of each selection term, we vary $\varepsilon$
in Equation~\ref{eq:selection} while keeping the top-$K$ shortlisting rule
($K=5$), a budget of $N=15$ candidate attempts, and other settings fixed.
With $\varepsilon=0$, parent sampling uses only the score-based softmax
distribution; with $\varepsilon=1$, it favors parents with fewer direct
children.
The default mixture ($\varepsilon=0.35$) achieves the highest mean test score
on all three benchmarks (Table~\ref{tab:epsilon}), averaging $79.33$ versus
$73.93$ for exploitation alone and $73.65$ for exploration alone, gains of
$5.40$ and $5.68$ percentage points, respectively. Improvements are larger on
TAU2-bench and SWE-bench Verified than on AppWorld. These results suggest
that the two selection signals are complementary: measured performance
directs refinement toward strong implementations, while expansion history
preserves opportunities to develop less-explored alternatives under a
limited evaluation budget.

\begin{table}[t]
\centering
\caption{Comparison with a stronger executor. All methods use GPT-5.4 as the executor. For \textsc{HarnessFlow} and Meta-Harness, we rerun harness search from the same initial harness using this executor. Cost is USD per test case at the executor's list price (\$2.50 per million input tokens and \$15.00 per million output tokens). Rows are ordered by average score.}
\label{tab:executor}
\small
\setlength{\tabcolsep}{3.5pt}
\renewcommand{\arraystretch}{1.15}
\begin{tabular}{@{}l cc cc cc cc@{}}
\toprule
& \multicolumn{2}{c}{\textbf{TAU2}} & \multicolumn{2}{c}{\textbf{AppWorld}} & \multicolumn{2}{c}{\textbf{WebShop}} & \multicolumn{2}{c}{\textbf{Avg.}} \\
\cmidrule(lr){2-3} \cmidrule(lr){4-5} \cmidrule(lr){6-7} \cmidrule(lr){8-9}
\textbf{Method} & \textbf{Score} & \textbf{Cost} & \textbf{Score} & \textbf{Cost} & \textbf{Score} & \textbf{Cost} & \textbf{Score} & \textbf{Cost} \\
\midrule
Vanilla & 83.76{\scriptsize$\pm$2.83} & \textbf{0.23} & 82.48{\scriptsize$\pm$1.09} & 0.38 & 56.29{\scriptsize$\pm$1.20} & \textbf{0.08} & 74.18 & 0.23 \\
Codex & 85.91{\scriptsize$\pm$1.59} & 0.47 & 75.52{\scriptsize$\pm$2.37} & 0.57 & \textbf{73.59{\scriptsize$\pm$0.89}} & 0.38 & 78.34 & 0.47 \\
Hermes & 85.81{\scriptsize$\pm$1.73} & 0.70 & 79.43{\scriptsize$\pm$1.36} & 0.62 & 72.76{\scriptsize$\pm$1.43} & 0.47 & 79.33 & 0.60 \\
OpenHands & 83.12{\scriptsize$\pm$3.64} & 0.48 & 83.71{\scriptsize$\pm$1.32} & 0.52 & 71.62{\scriptsize$\pm$0.59} & 0.41 & 79.48 & 0.47 \\
Claude Code & 82.69{\scriptsize$\pm$1.76} & 0.61 & 87.81{\scriptsize$\pm$1.92} & 0.81 & 72.06{\scriptsize$\pm$1.36} & 0.59 & 80.85 & 0.67 \\
\midrule
Meta-Harness & 84.59{\scriptsize$\pm$1.12} & 0.24 & 88.41{\scriptsize$\pm$0.73} & \textbf{0.20} & 72.12{\scriptsize$\pm$0.51} & 0.09 & 81.71 & \textbf{0.18} \\
\rowcolor{gray!15}
\textsc{HarnessFlow} & \textbf{88.06{\scriptsize$\pm$1.92}} & 0.31 & \textbf{90.95{\scriptsize$\pm$0.95}} & 0.25 & 71.72{\scriptsize$\pm$0.66} & 0.08 & \textbf{83.58} & 0.21 \\
\bottomrule
\end{tabular}
\end{table}

\Needspace{7\baselineskip}
\paragraph{Generalization to Other LLMs.}
With GPT-5.4~\citep{openai2026gpt54}, \textsc{HarnessFlow} achieves the highest three-benchmark
average of $83.58\%$, exceeding Claude Code and Meta-Harness by $2.73$ and
$1.87$ points, respectively (Table~\ref{tab:executor}), and leads on
TAU2-bench and AppWorld. Execution costs remain well below Claude Code
($\$0.21$ versus $\$0.67$ per task). Improvements thus persist with a
stronger executor.

\section{Related Work}
\label{sec:related}

\paragraph{Automated Harness Optimization.}
Automatic optimization spans several levels of the agent stack.
Prompt and LM-program optimizers use task feedback to improve instructions
and computation-graph
components~\citep{yang2024large,khattab2024dspy,yuksekgonul2024textgrad}.
Structural search extends the editable object to agent modules, graph
connectivity, and communication
topology~\citep{zhuge2024language,shang2025agentsquare,zhang2024g};
evolutionary multi-agent generation and adaptive architecture or model
routing further explore how agents are assembled and assigned
computation~\citep{yuan2025evoagent,zhang2025multi,yue2025masrouter}.
At the code level, ADAS generates executable agent designs using an archive
of prior discoveries, while AFlow combines MCTS with code-represented
workflow refinement~\citep{hu2025automated,zhang2025aflow}.
Within harness optimization, HARBOR searches configurations, PRISM targets
prompts and guarded tool-boundary middleware, and Self-Harness validates
failure-driven changes to an agent's operating
harness~\citep{sengupta2026harbor,ruan2026beyond,zhang2026self}.
Code-level methods include SICA, which edits its own implementation, and
Meta-Harness, whose proposer inspects archived code, scores, and
traces~\citep{robeyns2025self,lee2026meta}.
Agentic Harness Engineering, HarnessCompass, and AutoSaddler further
structure execution evidence, edit constraints, and validation to guide
harness
updates~\citep{lin2026agentic,zhang2026harnesscompass,park2026autosaddler}.
Archive- and lineage-based search also appears in the Huxley--G\"odel Machine
and DarwinX~\citep{wang2026huxley,zhang2026darwinx}, while
Hyperagents makes the improvement procedure itself
editable~\citep{zhang2026hyperagents}.
These studies establish both code-level refinement and explicit search over
retained designs. \textsc{HarnessFlow} builds on these directions by coupling
a fixed coding-agent proposer with MCTS-guided parent selection,
filesystem-grounded experience reuse, and reusable harness mechanisms to
allocate a limited candidate-evaluation budget across full-harness
refinement paths.
Further related work on harness engineering is discussed in
Appendix~\ref{app:related}.

\section{Conclusion}

\textsc{HarnessFlow} optimizes complete agent harnesses through MCTS-guided
search and coding-agent refinement. Across six benchmarks, it improves
average test scores over Vanilla and Meta-Harness by $10.89$ and $3.95$
percentage points, respectively, while running at a lower average execution
cost than every hand-engineered harness.

\paragraph{Limitations and Future Work.}
Two properties scope the present formulation. The search is guided by a
validation score, so it applies to tasks for which an automatic grader is
available, and it consumes evaluation compute, a one-time cost that is
amortized over subsequent deployment but that bounds how large a budget is
practical. \textsc{HarnessFlow} also organizes candidates as a refinement
tree, in which every harness descends from a single parent. A natural
extension is to search over a graph, so that a candidate can combine
mechanisms discovered on separate branches, which we leave to future work.

\subsection*{AI use statement}

In this work, we used generative AI tools to assist with limited code implementation
and editing, including code organization and implementation support.
We also used generative AI tools to improve the clarity and readability of the manuscript.
Generative AI tools were not used to formulate the research idea, develop the core methodology
or experimental design, derive theoretical claims or proofs, or interpret experimental results.
Other required-disclosure categories were not applicable to this work. All AI-assisted code was
reviewed and tested by the authors, and all AI-assisted writing was manually reviewed and revised.
The authors take full responsibility for the final content of this work, including its text, claims,
experimental results, code, and other artifacts.

\subsection*{Ethics statement}

This work uses publicly released benchmarks and involves no human subjects or
personal data. The harnesses produced by our search are executable programs;
every candidate is evaluated inside a container or sandbox with fixed time and
resource limits. We are not aware of other ethical concerns raised by this work.
The Claude Code baseline experiments reported in this paper were conducted by the
fourth author, Aidan Wu, while he was working in the United States.

\subsection*{Reproducibility Statement}

Appendix~\ref{app:setup} documents the data splits, initial harnesses, baseline
releases, and the search and model settings.
Our implementation and scripts for reproducing the tables and figures will
be released.

\clearpage
\bibliography{references}
\bibliographystyle{iclr2027_conference}

\clearpage
\appendix
\section{Additional Related Work}
\label{app:related}

\paragraph{Harness Engineering.}
LLM agents depend on runtime mechanisms that organize reasoning, retain
state, coordinate model calls, and mediate environment interaction.
Reasoning--acting loops, verbal feedback, and inference-time search provide
complementary ways to structure task
execution~\citep{yao2022react,madaan2023self,zhou2023language}.
Memory-centered architectures retain, summarize, and retrieve past
experiences to support long-horizon
behavior~\citep{park2023generative,packer2023memgpt}.
Multi-agent frameworks introduce role assignment, structured communication,
and coordinated
workflows~\citep{li2023camel,wu2023autogen,hong2024metagpt}.
Software-engineering agents further demonstrate the importance of tool
interfaces and execution
environments~\citep{yang2024swe,wang2025openhands}, while HarnessX organizes
composable components for adaptive harness
construction~\citep{chen2026harnessx}.
Together, these studies identify runtime mechanisms and their composition as
important determinants of agent behavior beyond the underlying model.
Our work takes the complete harness implementation as the unit of
optimization, allowing these interacting mechanisms to be refined jointly.
The closest search-based precedent is AFlow~\citep{zhang2025aflow}, which also
applies MCTS to code-represented candidates. It differs from our setting in
two ways. First, its search space is the
workflow: how a fixed vocabulary of operators is wired together, which is one
layer of a harness. The execution loop, state handling, tool interfaces, and
environment mediation stay fixed, and it is precisely these layers that the
mechanisms above, and the harnesses selected by our search, modify. Second,
its candidates are produced by a single model call that rewrites the current
workflow from a textual summary of past changes and scores, whereas our proposer is a coding agent that runs the harness,
reads its execution evidence, and makes coordinated edits across files. AFlow is accordingly evaluated on
single-turn question answering, mathematics, and code generation, whereas the
interactive benchmarks studied here are dominated by mechanisms that lie
outside the workflow layer.

\section{Harness Component Library}
\label{app:harness-library}

\begin{figure}[H]
\centering
\includegraphics[width=\textwidth]{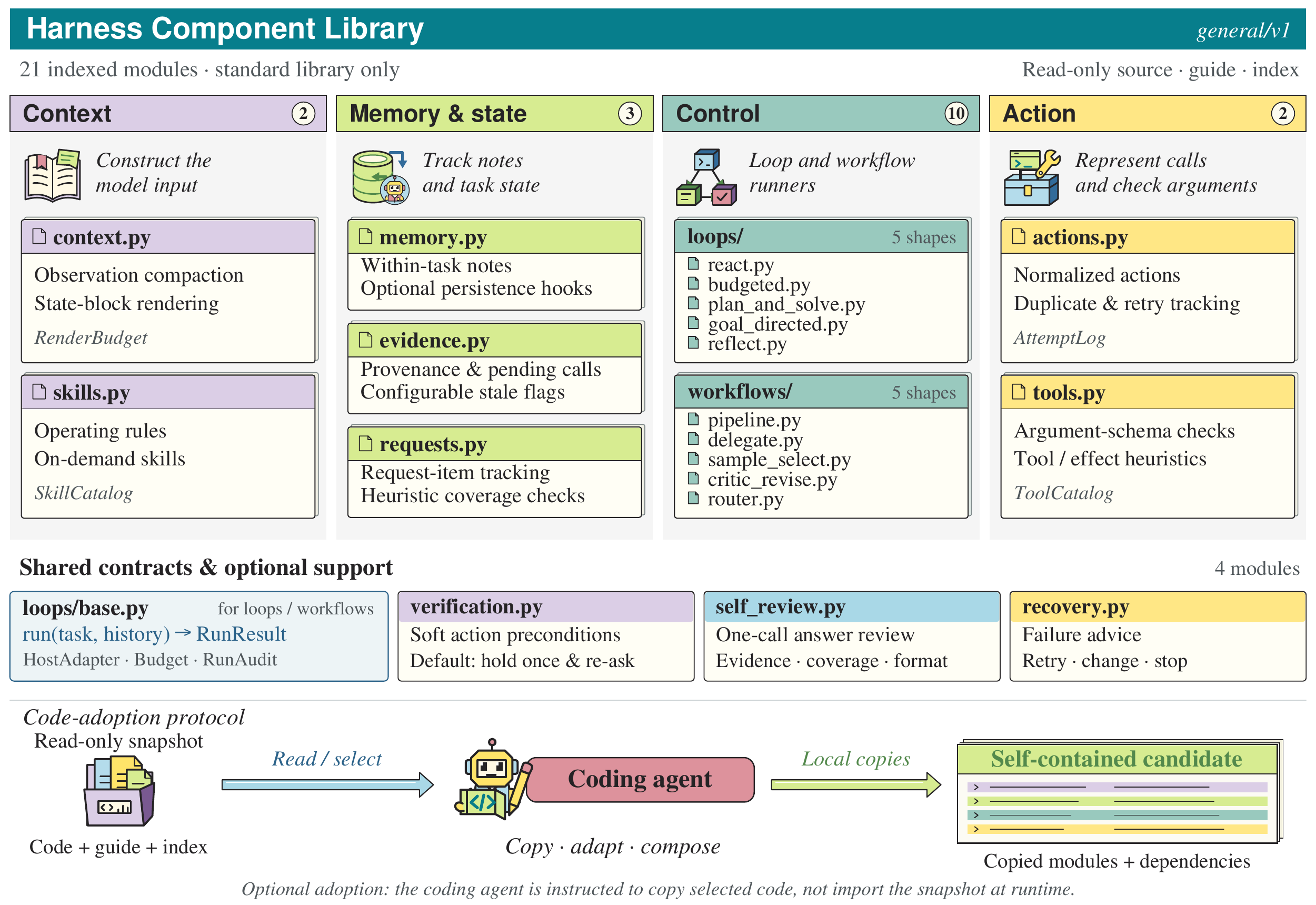}
\caption{Functional organization of the Harness Component Library's $21$
indexed modules. The four columns and the shared-support row indicate
functional roles. The coding agent copies adopted modules from a read-only
snapshot into the candidate.}
\label{fig:harness-library}
\end{figure}

Figure~\ref{fig:harness-library} gives a functional view of the library used
in our experiments. It contains $21$ benchmark-neutral Python modules
organized around four roles: context construction (observation compaction and
skill catalogs), memory and state tracking (note stores, observation logs, and
request tracking), control flow (loop and workflow runners that share a common
\texttt{run(task, history)} interface), and tool interfacing (argument-schema
checks and tool-effect classification), together with shared support for
verification, self-review, and recovery. The modules depend only on the
Python standard library; model calls and tool execution are supplied by the
host through a small adapter interface.
During refinement, the coding agent inspects a read-only snapshot of the
library and copies any adopted modules, with their local dependencies, into
the candidate, so that every candidate remains self-contained.

\section{Experimental Details}
\label{app:setup}

This appendix gives the data splits (\S\ref{app:data}), the initial harness
of each benchmark (\S\ref{app:harness0}), the baseline agents
(\S\ref{app:agents}), and the search and model settings shared by all methods
(\S\ref{app:config}).

\subsection{Benchmarks and Data}
\label{app:data}

Each benchmark is split once with a fixed random seed ($42$) into a validation
set used during search and a disjoint test set used only for the final
evaluation: $92$/$186$ tasks for TAU2-bench, $105$/$210$ for AppWorld, and
$100$/$200$ for WebShop, BrowseComp-Plus, SWE-bench Verified, and
BigCodeBench-Hard, whose tasks are sampled from the BigCodeBench release
using its published difficulty criteria. The same split and the benchmark's
official grader are used for every method.

\subsection{Initial Harnesses}
\label{app:harness0}

The initial harness of each benchmark (Vanilla in Table~\ref{tab:main}) is
the benchmark's own reference agent, reproduced from the upstream repository
at the revision listed in Table~\ref{tab:harness0}, with the model client
replaced by the shared gateway client so that the initial harness, every
search candidate, and every baseline call the same model.

\begin{table}[H]
\centering
\caption{The initial harness of each benchmark and its upstream source. Revision is the git commit (abbreviated to seven characters) or release of the upstream repository from which the harness was reproduced.}
\label{tab:harness0}
\small
\setlength{\tabcolsep}{6pt}
\renewcommand{\arraystretch}{1.15}
\begin{tabular}{@{}>{\raggedright\arraybackslash}p{3.3cm} >{\raggedright\arraybackslash}p{6.4cm} >{\raggedright\arraybackslash}p{3.0cm}@{}}
\toprule
\textbf{Benchmark} & \textbf{Reference agent} & \textbf{Revision} \\
\midrule
TAU2-bench & tau2-bench \texttt{LLMAgent}~\citep{barres2025tau} & \texttt{a2c0247} \\
AppWorld & Official ReAct experiment harness~\citep{trivedi2024appworld} & \texttt{66ad809} (v0.1.3.post1) \\
WebShop & ReAct WebShop scaffold~\citep{yao2022webshop} & \texttt{a5c0827} \\
BrowseComp-Plus & \texttt{search\_agent/openai\_client.py}~\citep{chen2026browsecomp} & \texttt{0469490} \\
SWE-bench Verified & mini-SWE-agent \texttt{DefaultAgent}~\citep{yang2024swe} & \texttt{04d809c} \\
BigCodeBench-Hard & \texttt{bigcodebench.generate}~\citep{zhuo2025bigcodebench} & v0.1.4 \\
\bottomrule
\end{tabular}
\end{table}

\subsection{Baseline Agents}
\label{app:agents}

The four production agents are run through their official non-interactive
entry points at the releases in Table~\ref{tab:agents}, with their default
tool sets, skills, and subagents enabled. All methods, candidates and
baselines alike, run in the same network-isolated sandbox whose only egress
is the shared model gateway, use the same backbone and per-benchmark limits
(Table~\ref{tab:config}), and access the same benchmark tools through a
benchmark-specific tool server. Each agent receives the benchmark's task
description together with its I/O conventions (tool names, answer format,
and submission protocol).

\begin{table}[H]
\centering
\caption{Production agent harnesses and the release used throughout.}
\label{tab:agents}
\small
\setlength{\tabcolsep}{8pt}
\renewcommand{\arraystretch}{1.15}
\begin{tabular}{@{}l l l@{}}
\toprule
\textbf{Agent} & \textbf{Release} & \textbf{Source} \\
\midrule
Claude Code~\citep{claudecode} & 2.1.272 & npm \texttt{@anthropic-ai/claude-code} \\
Codex~\citep{codexcli} & 0.154.0 & npm \texttt{@openai/codex} \\
OpenHands~\citep{wang2025openhands} & cloud-1.45.0 (SDK 1.34.0) & GitHub \\
Hermes~\citep{hermesagent} & v2026.9.7 & GitHub \\
\bottomrule
\end{tabular}
\end{table}

\subsection{Search and Model Configuration}
\label{app:config}

\paragraph{Search.}
Each search runs $N=15$ candidate attempts, and every proposed candidate
consumes one attempt. In Algorithm~\ref{alg:harnessflow}, $v_n$ is valid when
the candidate passes a runtime compatibility check and its evaluation yields
a score. Selection uses the parameters given in Section~\ref{sec:setup}, and
every candidate, including the root, is scored on the validation split with
$R=5$ repetitions per task. Meta-Harness is run through the same pipeline
with the same initial harness, coding agent, and candidate budget.

\paragraph{Coding agent.}
Candidates are produced by Codex~\citep{codexcli} (release $0.154.0$) running
GPT-5.5 at the highest reasoning effort. It receives the search tree of Section~\ref{sec:method} with its
scores and execution records, the Harness Component Library, and the
benchmark's task description. Meta-Harness uses the same coding agent, model, and
reasoning effort.

\paragraph{Models.}
All harnesses execute GPT-5.4-mini at medium reasoning effort through a local
gateway, with sampling temperature at the API default. Auxiliary models are
fixed per benchmark and shared by all methods: the TAU2-bench user simulator
is GPT-5.2 and its evaluator GPT-5.5, the BrowseComp-Plus grader is GPT-5.5
with the upstream grading template, and the remaining benchmarks use
programmatic graders. The GPT-5.4 experiments of Section~\ref{sec:ablation}
change only the executor model.

\begin{table}[H]
\centering
\caption{Per-benchmark execution limits, applied identically to every method. Deadlines are wall-clock per task.}
\label{tab:config}
\small
\setlength{\tabcolsep}{8pt}
\renewcommand{\arraystretch}{1.15}
\begin{tabular}{@{}l l l@{}}
\toprule
\textbf{Benchmark} & \textbf{Episode limit} & \textbf{Deadline} \\
\midrule
TAU2-bench & $200$ steps & $1{,}800$\,s \\
AppWorld & $50$ interactions & $1{,}800$\,s \\
WebShop & $30$ actions & $900$\,s \\
BrowseComp-Plus & $100$ tool calls & $1{,}800$\,s \\
SWE-bench Verified & $250$ model calls & $2{,}400$\,s \\
BigCodeBench-Hard & $30$ iterations & $1{,}800$\,s \\
\bottomrule
\end{tabular}
\end{table}

\section{Additional Results}
\label{app:results}

This appendix reports per-benchmark search curves (\S\ref{app:search}), a
second harness-evolution case (\S\ref{app:search-evolution}), and
per-benchmark execution costs (\S\ref{app:cost}).

\subsection{Search Behavior}
\label{app:search}

\begin{figure}[H]
\centering
\includegraphics[width=0.9\textwidth]{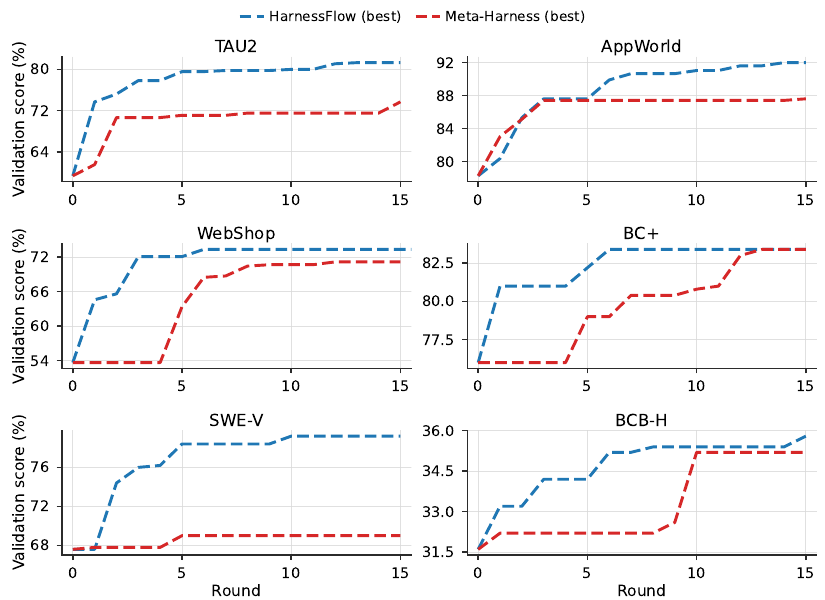}
\caption{Best validation score reached so far at each round, per benchmark, as the dashed lines of Figure~\ref{fig:curves}. Round $0$ is the shared initial harness.}
\label{fig:curves-round}
\end{figure}

Figure~\ref{fig:curves-round} shows, for every benchmark, the best validation
score reached so far at each round; Figure~\ref{fig:curves} in the main text
averages these curves over the six benchmarks.

\subsection{Harness Evolution on TAU2-bench}
\label{app:search-evolution}

Figure~\ref{fig:evolution-tau2} complements the SWE-bench Verified example of
Figure~\ref{fig:evolution-swe} with a tool-use benchmark, showing the
recorded parent relationships (a), per-round and best-so-far validation
scores (b), and the code refinements along the selected lineage (c).

\begingroup
\setlength{\intextsep}{7pt}
\captionsetup{font=small,skip=3pt}
\begin{figure}[H]
\centering
\includegraphics[width=\textwidth]{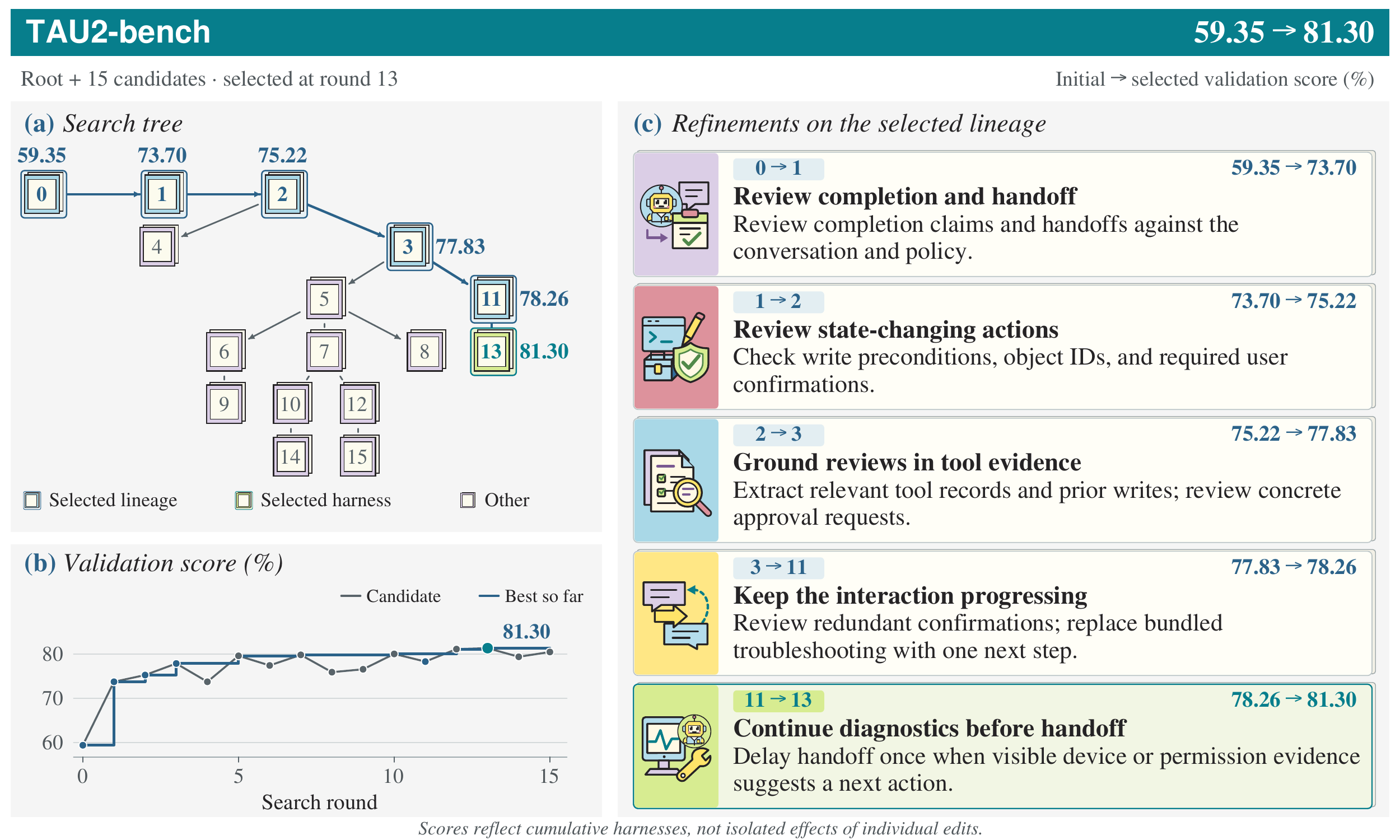}
\caption{Harness evolution on TAU2-bench. Action reviews, structured evidence,
and troubleshooting continuation raise the validation score from $59.35\%$ to
$81.30\%$.}
\label{fig:evolution-tau2}
\end{figure}
\endgroup

\subsection{Execution Cost}
\label{app:cost}

\begin{figure}[H]
\centering
\includegraphics[width=\textwidth]{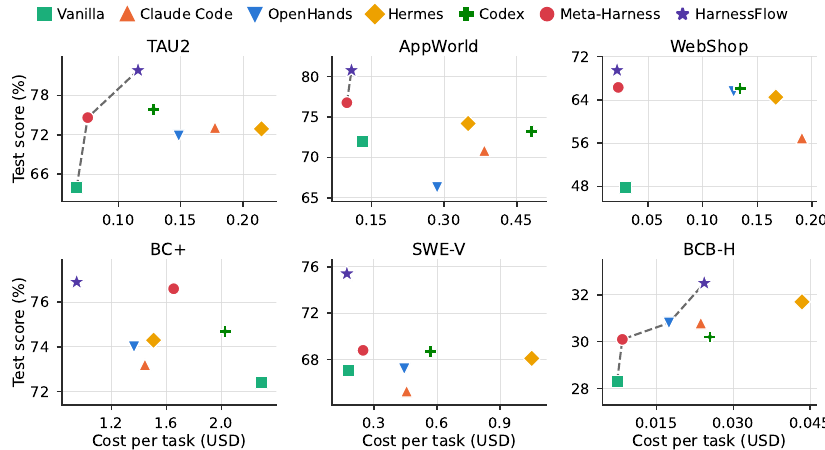}
\caption{Test score (Table~\ref{tab:main}) against execution cost per task (Table~\ref{tab:cost}) on each benchmark. The dashed line joins the methods on the Pareto frontier; axis limits differ per panel because costs span two orders of magnitude between benchmarks.}
\label{fig:pareto-bench}
\end{figure}

Table~\ref{tab:cost} lists the execution cost of every method on every
benchmark, and Figure~\ref{fig:pareto-bench} plots test score against cost
per benchmark; Figure~\ref{fig:pareto} in the main text shows the
six-benchmark averages. By arithmetic average, \textsc{HarnessFlow} has the
lowest cost at $\$0.232$ per task, against $\$0.398$--$\$0.561$ for the
hand-engineered harnesses, and it lies on the Pareto frontier of every
benchmark.

\begin{table}[H]
\centering
\caption{Execution cost of every method in Table~\ref{tab:main}, in USD per test task: input and output tokens priced separately at the executor's list price (GPT-5.4-mini: \$0.75 per million input tokens and \$4.50 per million output tokens). Bold marks the cheapest method per column.}
\label{tab:cost}
\small
\setlength{\tabcolsep}{5pt}
\renewcommand{\arraystretch}{1.15}
\begin{tabular}{@{}l cc cc cc c@{}}
\toprule
& \multicolumn{2}{c}{\textbf{Agentic}} & \multicolumn{2}{c}{\textbf{Web}} & \multicolumn{2}{c}{\textbf{Code}} & \\
\cmidrule(lr){2-3} \cmidrule(lr){4-5} \cmidrule(lr){6-7}
\textbf{Method} & \textbf{TAU2} & \textbf{AppWorld} & \textbf{WebShop} & \textbf{BC+} & \textbf{SWE-V} & \textbf{BCB-H} & \textbf{Avg.} \\
\midrule
Vanilla & \textbf{0.067} & 0.132 & 0.030 & 2.292 & 0.181 & \textbf{0.007} & 0.451 \\
Claude Code & 0.177 & 0.383 & 0.191 & 1.444 & 0.456 & 0.024 & 0.446 \\
OpenHands & 0.148 & 0.286 & 0.128 & 1.365 & 0.445 & 0.017 & 0.398 \\
Hermes & 0.215 & 0.350 & 0.167 & 1.507 & 1.050 & 0.043 & 0.555 \\
Codex & 0.128 & 0.481 & 0.134 & 2.027 & 0.569 & 0.025 & 0.561 \\
Meta-Harness & 0.075 & \textbf{0.100} & 0.023 & 1.653 & 0.249 & 0.008 & 0.351 \\
\rowcolor{gray!15}
\textsc{HarnessFlow} & 0.116 & 0.109 & \textbf{0.022} & \textbf{0.948} & \textbf{0.172} & 0.024 & \textbf{0.232} \\
\bottomrule
\end{tabular}
\end{table}

\end{document}